\documentclass[runningheads]{llncs}
\usepackage[T1]{fontenc}
\usepackage{graphicx}
\usepackage{hyperref}
\usepackage{amsmath,amsfonts}
\usepackage{mathtools}

\usepackage{algorithm}%
\usepackage{algorithmicx}%
\usepackage{algpseudocode}%
\usepackage{subfig}
\usepackage[labelfont=bf]{caption}
\usepackage{xcolor}
\usepackage{array}
\usepackage{multirow}
\usepackage{enumitem}
\usepackage{rotating}
\usepackage{caption}
\usepackage{tikz}
\usepackage{color}

\usepackage[section]{placeins}

\usepackage{amsmath,amsfonts,bm}

\def\eqref#1{equation~\ref{#1}}

\def\1{\bm{1}}

\DeclareMathAlphabet{\mathsfit}{\encodingdefault}{\sfdefault}{m}{sl}
\SetMathAlphabet{\mathsfit}{bold}{\encodingdefault}{\sfdefault}{bx}{n}

\newcommand{\R}{\mathbb{R}}

\begin{document}
\title{Uncertainty Propagation in XAI: A Comparison of Analytical and Empirical Estimators}
\titlerunning{Uncertainty Propagation in XAI}
%
\author{Teodor Chiaburu\inst{1}\orcidID{0009-0009-5336-2455} \and
Felix Bießmann\inst{1,2}\orcidID{0000-0002-3422-1026} \and
Frank Haußer\inst{1}\orcidID{0000-0002-8060-8897}
}
\authorrunning{T. Chiaburu et al.}
%
\institute{Berliner Hochschule für Technik, Berlin, Germany\\
\email{\{chiaburu.teodor;felix.biessmann;frank.hausser\}@bht-berlin.de} \and
Einstein Center Digital Future, Berlin, Germany}
\maketitle              
\begin{abstract}
Understanding uncertainty in Explainable AI (XAI) is crucial for building trust and ensuring reliable decision-making in Machine Learning models. This paper introduces a unified framework for quantifying and interpreting Uncertainty in XAI by defining a general explanation function $e_{\theta}(x, f)$ that captures the propagation of uncertainty from key sources: perturbations in  input data and model parameters. By using both analytical and empirical estimates of explanation variance, we provide a systematic means of assessing the impact uncertainty on  explanations. We illustrate the approach using a first-order uncertainty propagation as the analytical estimator. In a comprehensive evaluation across heterogeneous datasets, we compare analytical and empirical estimates of uncertainty propagation and evaluate their robustness. Extending previous work on inconsistencies in explanations, our experiments identify XAI methods that do not reliably capture and propagate uncertainty. 
Our findings underscore the importance of uncertainty-aware explanations in high-stakes applications and offer new insights into the limitations of current XAI methods. The code for the experiments can be found in our repository: \href{https://github.com/TeodorChiaburu/UXAI}{https://github.com/TeodorChiaburu/UXAI}.

\end{abstract}
\section{Introduction}\label{sec:intro}



Understanding uncertainty in Explainable AI (UXAI) is essential for assessing the reliability of explanations provided by Machine Learning (ML) models. Extensive research \cite{salcritique1,salcritique2,salcritique3,salcritique4} has shown that many widely used XAI methods produce explanations that are unreliable, non-robust and unfaithful to both the learned predictor and the data. These properties are widely recognized as fundamental desiderata of explanations in the XAI community \cite{ali_explainable_2023,quantus,schmidt2019quantifying}.

The absence of these properties can result in misleading attributions, which, in turn, undermines trust in model explanations. This issue is particularly problematic in tasks that require understanding how a model generalizes from data, such as debugging, feature importance analysis and identifying learned relationships between inputs and outputs. Ensuring reliability in XAI methods is especially critical in domains where human verification is difficult and explanation failures may go unnoticed, such as healthcare, finance or autonomous systems. Consequently, identifying and characterizing failure cases in XAI methods is a necessary step towards developing explanations that remain reliable across different inputs, model configurations and data distributions.

Despite these concerns, the degree to which explanations are sensitive to perturbations in the input and model parameters remains rather underexplored. If explanations vary significantly under small changes to the data or the model, this raises concerns about their stability and robustness in real-world applications. To address this gap, we introduce a principled framework for quantifying and analyzing uncertainty in explanations, focusing on how perturbations propagate through explanation functions. 
The main contributions of our work are:
\begin{itemize}
    \item[1)] We systematically compare two fundamental approaches to uncertainty quantification in XAI: \textbf{empirical estimation via Monte Carlo sampling} and \textbf{analytical estimation via first-order uncertainty propagation}. As a simplified measure of uncertainty, we compute the total variance of the explanations, given by the trace of their covariance matrices. 
    \item[2)] We experimentally investigate the conditions under which Gaussian perturbations in the input and model parameters lead to an approximately Gaussian distribution of explanations, highlighting the regimes where uncertainty propagates in a predictable manner across different XAI methods.
\end{itemize}

\section{Related Work}\label{sec:relwork}

Previous work on uncertainty in XAI (UXAI) has been focusing on developing explanation methods that inherently incorporate uncertainty estimation \cite{uxai_review}. 
They are designed to generate multiple explanations for the same input and predictor, allowing for direct uncertainty quantification through variance estimation. In \cite{uxai_review}, they are referred to as \textit{stochastic XAI methods}. Here, we further differentiate between \textit{parametric stochastic} (some of the parameters in $\theta$ are random variables such as randomly generated perturbations \cite{salmaps4,kernelshap}) and \textit{parametric deterministic} (the method admits parameters $\theta$, but they are not random variables \cite{occlusion}).

Either way, their variability comes from a set of XAI-method-specific parameters $\theta$. For instance, \cite{tcav,ace,ibd,copronn} compute concept scores via multiple random partitions of a 'non-concept' dataset. \cite{schwab2019cxplain} train an ensemble of UNets on previously generated explanations (from various XAI methods) and use the resulting distribution of explanations at inference time to measure uncertainty in the feature importance estimates. \cite{slack_reliable_2021,zhao_baylime_2021} use LIME \cite{salmaps4} as a backbone XAI method and estimate a confidence interval around the relevance scores of the features.

While these approaches enable direct uncertainty assessment within the explanation process, they are limited to XAI methods that are explicitly designed to be parameterized. This leaves open the question of how uncertainty propagates through non-parametric explanation methods, that provide point estimates of feature importance but do not naturally offer a measure of uncertainty. Some limited research looked into how sensitive explanations are with respect to perturbations in the input or the predictor's weights.

The authors of \cite{salcritique2} investigated the reliability of saliency-based explanation methods by examining their sensitivity to transformations that do not affect model predictions. Specifically, it has been shown that many saliency methods fail to satisfy \textit{input invariance}, meaning they incorrectly attribute importance when a constant shift is added to the input, even though this transformation has no effect on the model’s output. \cite{salcritique4} examine several methods' sensitivity to model parameters and the data-generating process. They introduce a \textit{randomization-based evaluation framework}, demonstrating that certain widely used saliency methods produce similar explanations regardless of whether the model is trained or randomly initialized. 


In this work, we focus on the propagation of uncertainty from the model input and model parameters to the explanation itself. This perspective is, to the best of our knowledge, largely unexplored in existing literature.
By leveraging both empirical and analytical first order approximations, we aim to bridge this gap and provide a structured methodology for measuring UXAI.
Our framework offers insights into how sensitive explanations are to perturbations in the input and model parameters, enabling a deeper understanding of the stability and reliability of widely used XAI techniques.

Our study primarily investigates uncertainty propagation in XAI by introducing controlled perturbations to inputs and model parameters. This analysis relates to the classical distinction between \textit{epistemic and aleatoric uncertainty}. Epistemic uncertainty arises from limited knowledge about the model parameters due to finite training data, while aleatoric uncertainty stems from inherent randomness in the data itself. In our framework, model weight perturbations can be viewed as approximating epistemic uncertainty, as they simulate the variability that would result from training different models on the same dataset. Input perturbations, on the other hand, capture aspects of aleatoric uncertainty by modeling variations in the observed data.

\section{Methods}\label{sec:methods}
This section presents the two considered sources of propagating uncertainty over to explanations - input perturbations and model weights perturbations - as well as approaches to quantify these propagations. 

\subsection{Formalizing Uncertainty Propagation in XAI}\label{subsec:formal_uxai}
We are considering various explanation methods $e$ for a number of prediction models $f$ (classification or regression). Similar to the notations in \cite{uxai_review}, let the explainer be a function
$$e_{\theta}: \mathbb{R}^{n \cdot l \cdot r} \rightarrow \mathbb{R}^m , $$
which maps an input sample $x$ (be it image, text or tabular data) to an explanation $e_{\theta}(x, f)$, where:
\\
\begin{enumerate}
    \item $n$ is the dimension of the input $x = (x_1, ..., x_n)^T$
    \item $l$ is the number of the parameters of the XAI method $\theta = (\theta_1, ..., \theta_l)^T$ (if applicable)
    \item $r$ denotes the number of the model parameters $\omega = (\omega_1, ..., \omega_r)^T$ of the prediction model $f := f(\omega, \cdot): \, x \mapsto f(\omega, x)$
    \item $m$ is the dimension of the explanation $e_{\theta}(x, f) = (e_1, ..., e_m)^T$ .
\end{enumerate}

To systematically assess the sensitivity of explanations, we analyze the Jacobian of the explainer, which encodes how small perturbations in the input $x$, the explainer's internal parameters $\theta$ and the predictor model $f$ influence the resulting explanation\footnote{For readability, we leave out the parameter list in the Jacobian and denote $\mathbb{J}_e(x,f)$ by $\mathbb{J}_e$. The same applies for its partial derivatives.}:
\begin{equation}
\label{eq:Jac_all}
\mathbb{J}_e = \bigg[\mathbb{J}_{e,x} \quad \mathbb{J}_{e,\theta} \quad \mathbb{J}_{e,\omega}\bigg] =  \bigg[\frac{\partial e}{\partial x} \quad \frac{\partial e}{\partial \theta} \quad \frac{\partial e}{\partial \omega}\bigg] \in \R^{m \times (n \cdot l \cdot r)} 
\end{equation}
Note that, for simplification reasons, we will only look at artificial neural networks as $f$ and the impact of their weights $\omega$ in $f(\omega)$ on the explainer $e$, while keeping in mind that there are more model variables that influence the computation of $e_{\theta}(x,f)$, e.g. the number of layers.

As for the parameter $\theta$: it is only relevant for parametric XAI methods, e.g. Occlusion \cite{occlusion} which controls the patch size, patch stride and occlusion value as parameters or CRAFT \cite{fel2023craft}, also parameterized with a patch size for extracting concepts. This paper will focus only on the influence of the uncertainty in the input $x$ and the uncertainty of predictor $f$ on the explainer $e$, so let henceforth $e_{\theta}(x,f)$ be simply $e(x,f)$ without loss of generality.
\subsection{Propagating the Uncertainty of Input Variables}
If we look only at the sensitivity of the explanation w.r.t. to the input $x$, then the derivative in \autoref{eq:Jac_all} is given by the Jacobian block:
\begin{equation}
\mathbb{J}_{e,x} \in \R^{m \times n} , \quad
\big( \mathbb{J}_{e,x} \big)_{i,j} = \cfrac{\partial e_i }{ \partial x_j} .
\end{equation}
In the following, we consider normally distributed perturbations  in the input $x$:
\begin{equation}
    \widetilde{x} = x + \Delta x, \quad 
    \Delta x \sim \mathcal{N}(0, \Sigma_{\Delta x}) ,
\end{equation}
where $\Sigma_{\Delta x} \in \mathbb{R}^{n \times n}$ is the covariance matrix of the added noise. By linearizing the perturbed explainer around $x$ we arrive at:
\begin{equation}
\label{eq:e_lin_x}
    e(\widetilde{x},f) \approx e(x,f) + \mathbb{J}_{e,x} \cdot \Delta x .
\end{equation}
Since the covariance of an affine transformation $ y = Ax  + b$ of a random vector $x$ is given by $\mathrm{Var}(y) = \mathrm{Var}(Ax + b) = A \cdot \mathrm{Var}(x) \cdot A^T$, the first order approximation of the covariance of the explanations $e$ is obtained as 
\begin{equation}\label{eq:var_lin}
    \mathrm{Var}\big(e(\widetilde{x},f)\big) \approx \mathrm{Var}\big(e(x,f) + \mathbb{J}_{e,x} \cdot \Delta x\big)  = \mathbb{J}_{e,x} \cdot \Sigma_{\Delta x} \cdot \mathbb{J}_{e,x}^T .
\end{equation}
%
Moreover, 
$$
e_{\text{lin},x} := e(x,f) + \mathbb{J}_{e,x} \cdot \Delta x
$$ 
is also normally distributed \footnote{\href{https://statproofbook.github.io/P/mvn-ltt.html}{https://statproofbook.github.io/P/mvn-ltt.html}} with
\begin{equation}\label{eq:e_lin_normal}
    e_{\text{lin},x} \sim \mathcal{N}\big(e(x,f), \mathbb{J}_{e,x} \cdot \Sigma_{\Delta x} \cdot \mathbb{J}_{e,x}^T\big)
\end{equation}

If the perturbation in the input features is normal and i.i.d., with feature variance $\sigma^2$, then the covariance of the input perturbation becomes diagonal:
\begin{equation}\label{eq:same_var_input}
\Sigma_{\Delta x} = \sigma^2 I, \quad 
\Delta x \sim \mathcal{N}(0, \sigma^2 I), 
\end{equation}
with 
$I$ denoting the $n \times n$ identity matrix.
Hence, in this case, the first order approximation of the variance is given by:
\begin{equation}\label{eq:cov_approx_input}
\mathrm{Var}(e(\widetilde{x},f)) \approx \mathrm{Var}(e_{\text{lin},x}) = \sigma^2 \cdot \mathbb{J}_{e,x} \cdot \mathbb{J}_{e,x}^T
\end{equation}

\subsection{Propagating the Uncertainty of Model Parameters}
If looking only at the sensitivity of the explanation w.r.t. to the weights $\omega$ of the predictor model $f$, then the derivative is:

\begin{equation}
\mathbb{J}_{e,\omega}\in \R^{m \times r} , \quad
\big( \mathbb{J}_{e,\omega} \big)_{i,j} = \cfrac{\partial e_i }{ \partial \omega_j}.
\end{equation}
%
We similarly perturb the weights $\omega$:
\begin{equation}
    \widetilde{\omega} = \omega + \Delta \omega, \quad \Delta \omega \sim \mathcal{N}(0, \Sigma_{\Delta \omega}),
\end{equation}
with $\Sigma_{\Delta \omega} \in \mathbb{R}^{r \times r}$.
By analogy to \ref{eq:var_lin} and \ref{eq:same_var_input}, considering i.i.d. perturbations with variances $\sigma^2$, the first order approximation of the uncertainty in the explanation w.r.t. the model weights $\omega$ is given by:
\begin{equation}\label{eq:cov_approx_model}
\mathrm{Var}\big( e(x, \widetilde{f}) \big) \approx \sigma^2 \cdot \mathbb{J}_{e,\omega} \cdot \mathbb{J}_{e,\omega}^T,
\end{equation}
where $\widetilde{f} = f(\widetilde{\omega})$.

\subsection{Quantifying Uncertainty in XAI}

We will quantify the uncertainty of an explanation $e(x,f)$ in terms of the variance-covariance matrix (in the following termed covariance matrix) $\Sigma_e = \mathrm{Var}(e) \in \mathbb{R}^{m \times m}$ of the corresponding multivariate distribution of explanations, which is obtained by propagating uncertainty of the input $x$ and/or model $f$.

 As a scalar metric across different explanation methods as well as prediction models, the trace  $\mathrm{Tr}(\Sigma_e)$ of the covariance matrix is used, divided by the squared norm of the reference explanation $e(x,f)$ and the explanation's dimension $m$.
 We call this scalar proxy for uncertainty quantification the \textbf{Mean Uncertainty in the Explanation (MUE)}:
 \begin{equation}\label{eq:mue}
    \mathrm{MUE} = \cfrac{\text{Tr}(\Sigma_e)}{m \, ||e(x,f)||_2^2}
\end{equation}
 Due to similarity-invariance, the trace of $\Sigma_e$ equals the sum of its eigenvalues \cite{strang2006linear} and is a well-established measure of the total variance in multidimensional distributions\footnote{\href{https://rpubs.com/mpfoley73/496132}{https://rpubs.com/mpfoley73/496132}}.
 
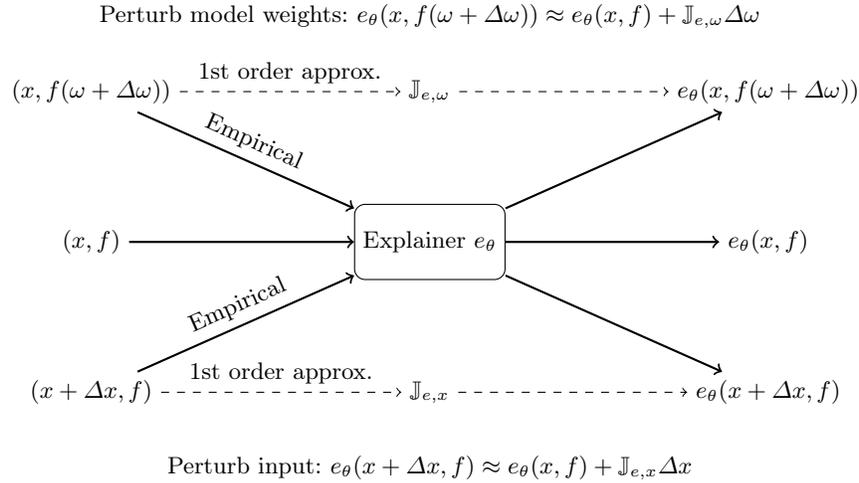
\begin{figure}[h]
    \centering
    \begin{tikzpicture}[node distance=2.5cm, auto]
    
        \node (xf) at (-1.5,0) {$(x, f)$};
        \node (perturb_x) at (-1.5,-2) {$(x + \Delta x, f)$};
        \node (perturb_omega) at (-1.5,2) {$(x, f(\omega + \Delta \omega))$};
        \node [draw, rectangle, rounded corners, minimum width=2cm, minimum height=1cm] (explainer) at (3,0) {Explainer $e_{\theta}$};
        \node (explanation) at (7.5,0) {$e_{\theta}(x, f)$};
        \node (perturbed_exp_x) at (7.5,-2) {$e_{\theta}(x + \Delta x, f)$};
        \node (perturbed_exp_omega) at (7.5,2) {$e_{\theta}(x, f(\omega + \Delta \omega))$};
        \node (jacobian_x) at (3,-2) {$\mathbb{J}_{e,x}$};
        \node (jacobian_omega) at (3,2) {$\mathbb{J}_{e,\omega}$};
        
        \draw[->, thick] (xf) -- (explainer);
        \draw[->, thick] (explainer) -- (explanation);
        \draw[->, thick] (perturb_x) -- (explainer) node[midway, above, sloped] {\footnotesize Empirical};
        \draw[->, thick] (explainer) -- (perturbed_exp_x);
        \draw[->, dashed] (perturb_x) -- (jacobian_x) node[midway, above, sloped] {\footnotesize 1st order approx.};
        \draw[->, dashed] (jacobian_x) -- (perturbed_exp_x);
        \draw[->, thick] (perturb_omega) -- (explainer) node[midway, above, sloped] {\footnotesize Empirical};
        \draw[->, thick] (explainer) -- (perturbed_exp_omega);
        \draw[->, dashed] (perturb_omega) -- (jacobian_omega) node[midway, above, sloped] {\footnotesize 1st order approx.};
        \draw[->, dashed] (jacobian_omega) -- (perturbed_exp_omega);
        
        \node at (3,-3) {\footnotesize Perturb input: $e_{\theta}(x + \Delta x, f) \approx e_{\theta}(x, f) + \mathbb{J}_{e,x} \Delta x$};
        \node at (3,3) {\footnotesize Perturb model weights: $e_{\theta}(x, f(\omega + \Delta \omega)) \approx e_{\theta}(x, f) + \mathbb{J}_{e,\omega} \Delta \omega$};
    
    \end{tikzpicture}
    \caption{Empirical Monte Carlo and first order approximation approaches for estimating the propagation of uncertainty in the input $x$ (bottom) and model weights $\omega$ (top) to the explanation $e_{\theta}$. 
    }
    \label{fig:jac_lin}
\end{figure}

We consider two approaches for computing approximations of the covariance matrix $\Sigma_e$, see \autoref{fig:jac_lin}: 
\begin{itemize}
    \item \textbf{Monte Carlo simulations} lead to an \textit{empirical covariance matrix} $\Sigma_\mathrm{MC}$, derived from multiple explanations obtained with perturbed input or perturbed model parameters.
    \item \textbf{First-order approximations} based on local sensitivity analysis as presented in \ref{eq:cov_approx_input} and \ref{eq:cov_approx_model} lead to an \textit{analytical covariance matrix} $\Sigma_\mathrm{lin}$. Here, no sampling is needed and the partial derivatives are computed using finite differences (see repository\footnote{One could optimize the computation of the derivatives further using automatic differentiation.}).
\end{itemize}

\autoref{alg:inp_pert} in the Supplement describes these computations in detail, exemplarily for the case of uncertainty propagation from input $x$ to explainer $e$. The propagation stemming from the model weights $\omega$ is analyzed analogously.

By comparing the Mean Uncertainty in the Explanation (MUE) for both methods, i.e. using $\Sigma_\mathrm{MC}$ or $\Sigma_\mathrm{lin}$,
we evaluate the consistency of the two approaches and highlight potential discrepancies in uncertainty quantification across different XAI methods.

\section{Design of Experiments}\label{sec:experiments}
For our experiments we used two well-known ML tasks from the literature: classification on MNIST Handwritten Digits \footnote{\href{http://yann.lecun.com/exdb/mnist/}{http://yann.lecun.com/exdb/mnist/}} and regression on Auto MPG \footnote{\href{https://archive.ics.uci.edu/dataset/9/auto+mpg}{https://archive.ics.uci.edu/dataset/9/auto+mpg}}. We trained a simple CNN classifier on the first task and a simple MLP regressor on the second one. 

For computing the explanations described in the next section, we used the \textit{xplique} library \cite{xplique}. We looked at five standard feature attribution methods: four gradient based (Saliency \cite{grad}, GradientInput \cite{gradientinput}, GuidedBackprop \cite{guidedbp}, Integrated Gradients \cite{integ}) and one perturbation based (Occlusion \cite{occlusion}). Note, however, that our approach can also be applied to other classes of explainers, e.g. concept-based (where the explanation vector $(e_1, ..., e_m)^T$ contains concept scores/similarities) or example-based (where the explanation vector contains the rankings or again similarities to the relevant samples).

To investigate how uncertainty propagates in XAI methods, we compare \textbf{empirical uncertainties} against \textbf{first-order approximations} for the aforementioned XAI-methods and for the two test cases. As detailed in \autoref{sec:methods}, we consider perturbations in the input $x$ or in the model parameters $\omega$. Specifically, we examine how uncertainties in explanations behave as a function of the perturbation scale $\sigma^2$. The values considered for $\sigma$ are $\{10^{-6}, 10^{-5}, 10^{-4}, 10^{-3}, 10^{-2}, 10^{-1}\}$ (we refer to this as the \textit{low variance regime}) and $\{0.2, 0.3, 0.4, 0.5\}$ (the \textit{high variance regime}).

For perturbing the input $x$, we added random noise $\Delta x$ to the whole input vector $x$ (every RGB-pixel in the case of the $28\times28\times3$-images\footnote{While the original MNIST images are gray-scale, we converted them into RGB, so that we can test a broader range of XAI methods from the \textit{xplique} library, that require 3 color channels.} in MNIST and each of the 9 tabular features in the case of Auto MPG). As far as perturbations in $\omega$ are concerned, we restricted ourselves to only perturbing the weights (without the bias) in the final Dense layer of both the CNN for classification and the MLP for regression; this accounts, then, for 64 scalar weights in the MLP and 64 $\cdot$ 10 in the CNN (because of the 10 digit classes).

For orientation, the partial derivatives (Jacobian blocks) used for the linear approximations have the following dimensions:

\begin{itemize}
    \item MNIST: $\mathbb{J}_{e,x} \in \mathbb{R}^{(28 \cdot 28) \times (28 \cdot 28 \cdot 3)} \quad \text{and} \quad \mathbb{J}_{e,{\omega}} \in \mathbb{R}^{(28 \cdot 28) \times (64 \cdot 10)}$
    \item Auto MPG: $\mathbb{J}_{e,x} \in \mathbb{R}^{9 \times 9} \quad \text{and} \quad \mathbb{J}_{e,{\omega}} \in \mathbb{R}^{9 \times 64}$
\end{itemize}

We refer the reader to the visualization examples of the Jacobian blocks and of the covariance matrices in \autoref{fig:diagonals} in the Supplement, as well as in our repository.

In both tasks, samples were randomly chosen from their datasets, for which the corresponding values of the Mean Uncertainty in Explanation (MUE), as defined in \autoref{eq:mue}, were computed and averaged for each value of the perturbation parameter $\sigma^2$. Interestingly, we observe that some XAI methods do not reliably propagate uncertainties. 

\section{Results} \label{sec:results}
\begin{figure}[h]
    \centering
    \subfloat[Auto MPG (regression)]{\includegraphics[width=1.0\textwidth]{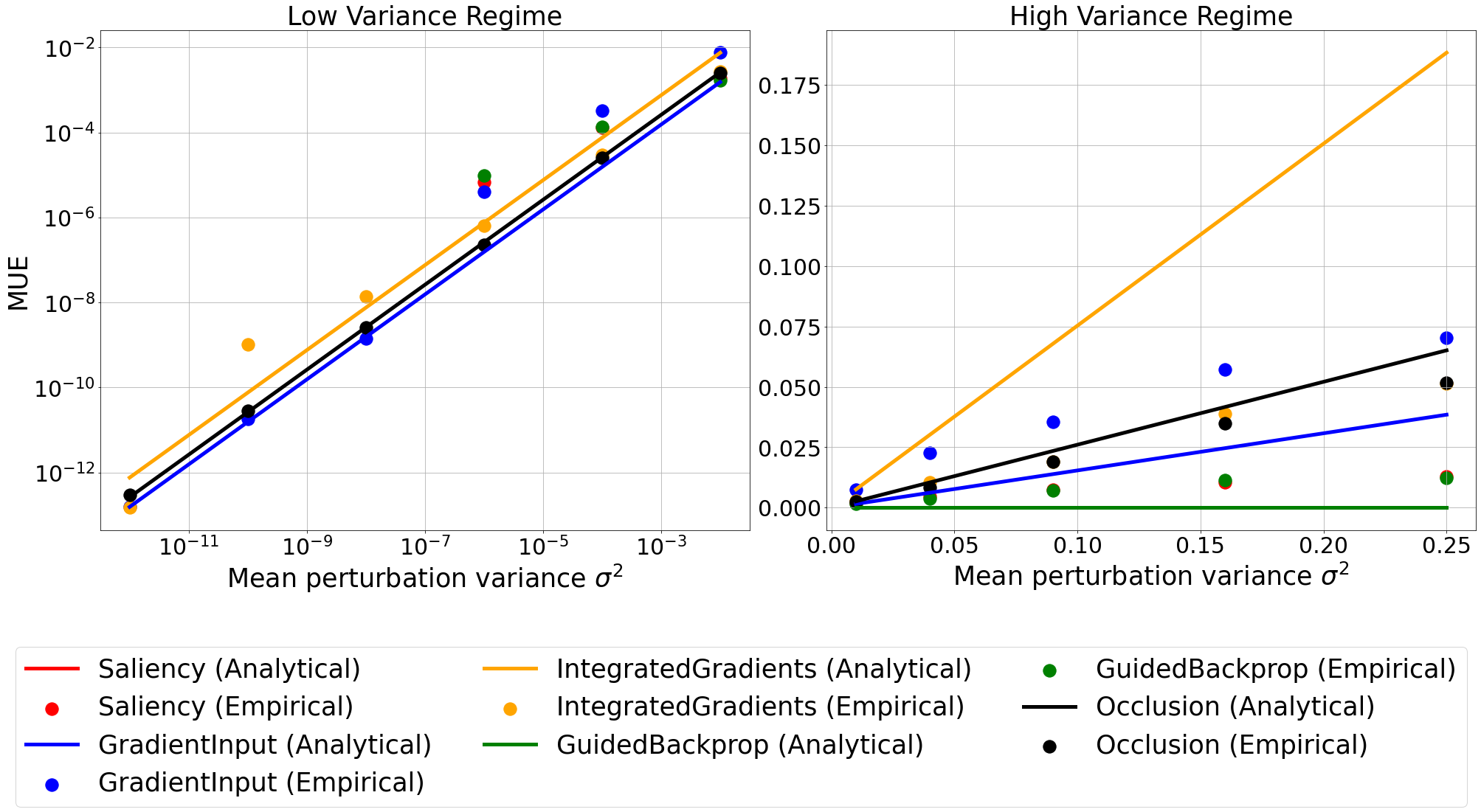}}
    \hfill
    \subfloat[MNIST (classification)]{\includegraphics[width=1.0\textwidth]{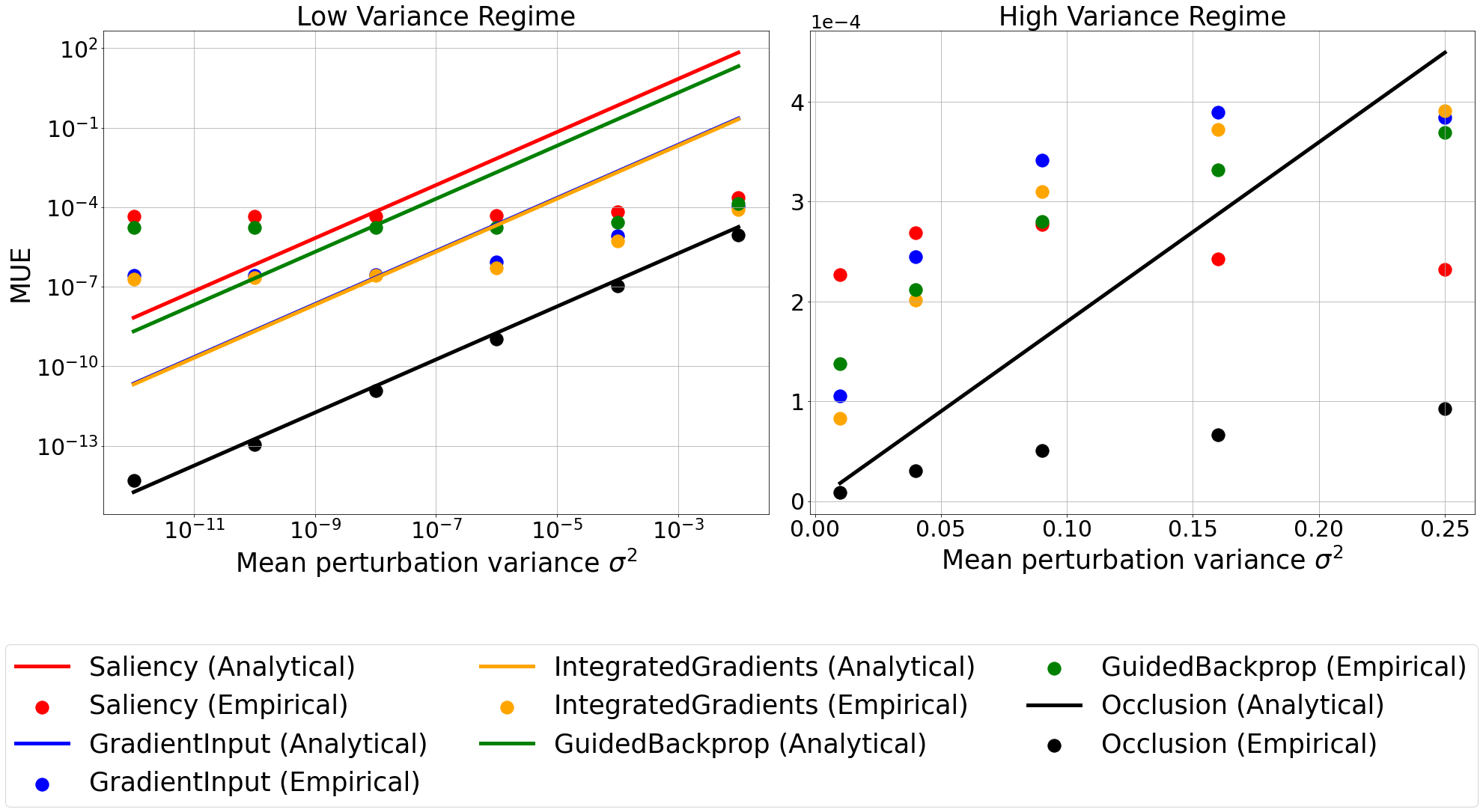}}
    \hfill
    \caption{Impact of uncertainty in input variables on uncertainty of explanations. Gradient-based XAI methods exhibit a plateau in the empirical MUE on MNIST - Case 3, \textbf{(b) left}. Saliency and GuidedBackprop have a MUE of 0 on the regression task, when linearized - Case 2, \textbf{(a)}. Analytical forecasts for MUE align with the empirical results for Occlusion on both datasets - Case 1, \textbf{(a, b)}. All results are aggregated over 10 random samples from both datasets. The continuous lines show the MUE forecast computed from $\Sigma_{\text{lin}}$, while the dots mark the MUE computed from $\Sigma_{\text{MC}}$. Please beware of the different axes scales (logarithmic on the left, linear on the right). Lines not visible in the high variance regime plots are left out because of their high slope. More individual results can be seen in \autoref{fig:uncert_input_ex_tab} and \autoref{fig:uncert_input_ex_mnist} in the Appendix.
    }
    \label{fig:uncert_input_agg}
\end{figure}

\begin{figure}[h]
    \centering
    \subfloat[Auto MPG (regression)]{\includegraphics[width=1.0\textwidth]{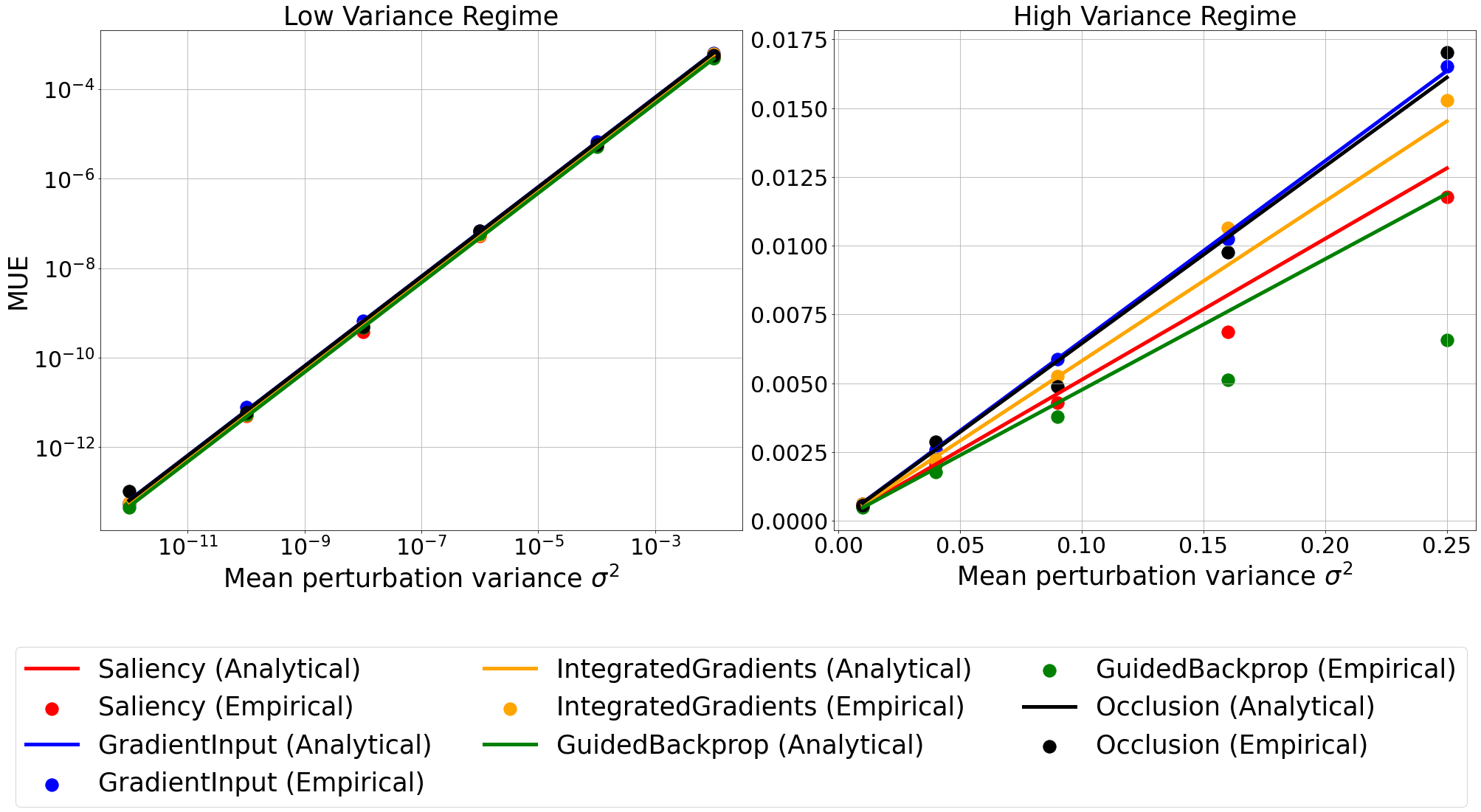}}
    \hfill
    \subfloat[MNIST (classification)]{\includegraphics[width=1.0\textwidth]{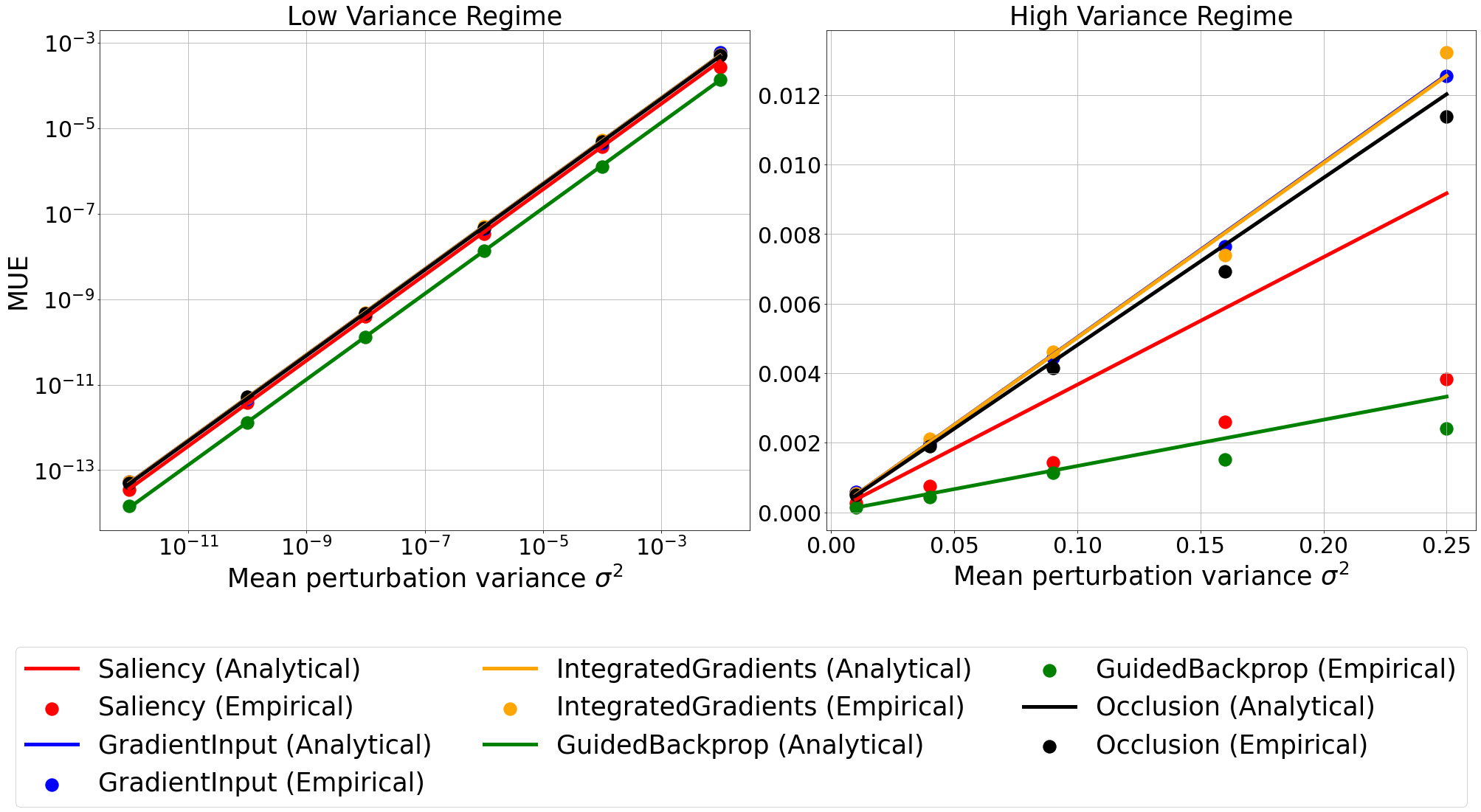}}
    \hfill
    \caption{Impact of uncertainty in model parameters on uncertainty of explanations. Empirical MUE w.r.t. perturbations in $\omega$ always follows linearization forecasts, across all tested XAI methods, all $\sigma^2$ regimes and datasets - Case 1. All results are aggregated over 10 random samples from both datasets. The continuous lines show the MUE forecast computed from $\Sigma_{\text{lin}}$, while the dots mark the MUE computed from $\Sigma_{\text{MC}}$. Please beware of the different axes scales (logarithmic on the left, linear on the right). More individual results can be seen in \autoref{fig:uncert_weights_ex_tab} and \autoref{fig:uncert_weights_ex_mnist} in the Appendix.}
    \label{fig:uncert_weights_agg}
\end{figure}

Our main results for uncertainty propagation of input data uncertainty are presented in \autoref{fig:uncert_input_agg} and for uncertainty propagation of model uncertainty  in \autoref{fig:uncert_weights_agg}.
We identified three distinct scenarios, which will be described in the following subsections. 
%
%
\subsection{Case 1: Analytical and Empirical Estimates Align}
In this scenario, empirical explanation uncertainties increase proportionally to the variance of input or model uncertainty $\sigma^2$, aligning well with the first-order approximations of analytical estimates of uncertainty propagation. 
This case describes particularly well the effect of perturbations in model parameters $\omega$ on the explanation - \autoref{fig:uncert_weights_agg}. There, the MUE predicted by the first order approximations aligns very well with the empirically computed MUE. We also observe this behavior for perturbations in input variables $x$, however, only for explanations computed with Occlusion (on both tasks) or with GradientInput and Integrated Gradients (on the tabular dataset) as shown in  \autoref{fig:uncert_input_agg}. The distributions in the explanations shown in \autoref{fig:input_normality} and \autoref{fig:weights_normality} show that there is a correlation between Case 1 and how well the variance in the perturbed explanations scales with $\sigma^2$.
%
%
\subsection{Case 2: Near-Zero Empirical Uncertainty for Small $\sigma^2$}
Certain methods exhibit a zero-threshold effect, where empirical uncertainties remain close to zero for small perturbations, while the linearization forecasts are also frozen at zero - \autoref{fig:uncert_input_agg}. This occurs for Saliency and GuidedBackProp when applied to the tabular dataset. The distributions in the explanations shown in \autoref{fig:input_normality} also report no variance in the explanations computed with these methods on Auto MPG. The likely cause is that these methods rely on gradient computations, and given that the regressor is a simple MLP with ReLU activations as nonlinearity, the gradient is constant for small perturbations leading to constant explanations in the MC-simulations and to a zero Jacobian block for the linear approximation.
\subsection{Case 3: Uncertainty Plateaus Below a Certain $\sigma^2$}
\begin{figure}
    \centering
    \subfloat[Auto MPG (regression)]{\includegraphics[width=1.0\textwidth]{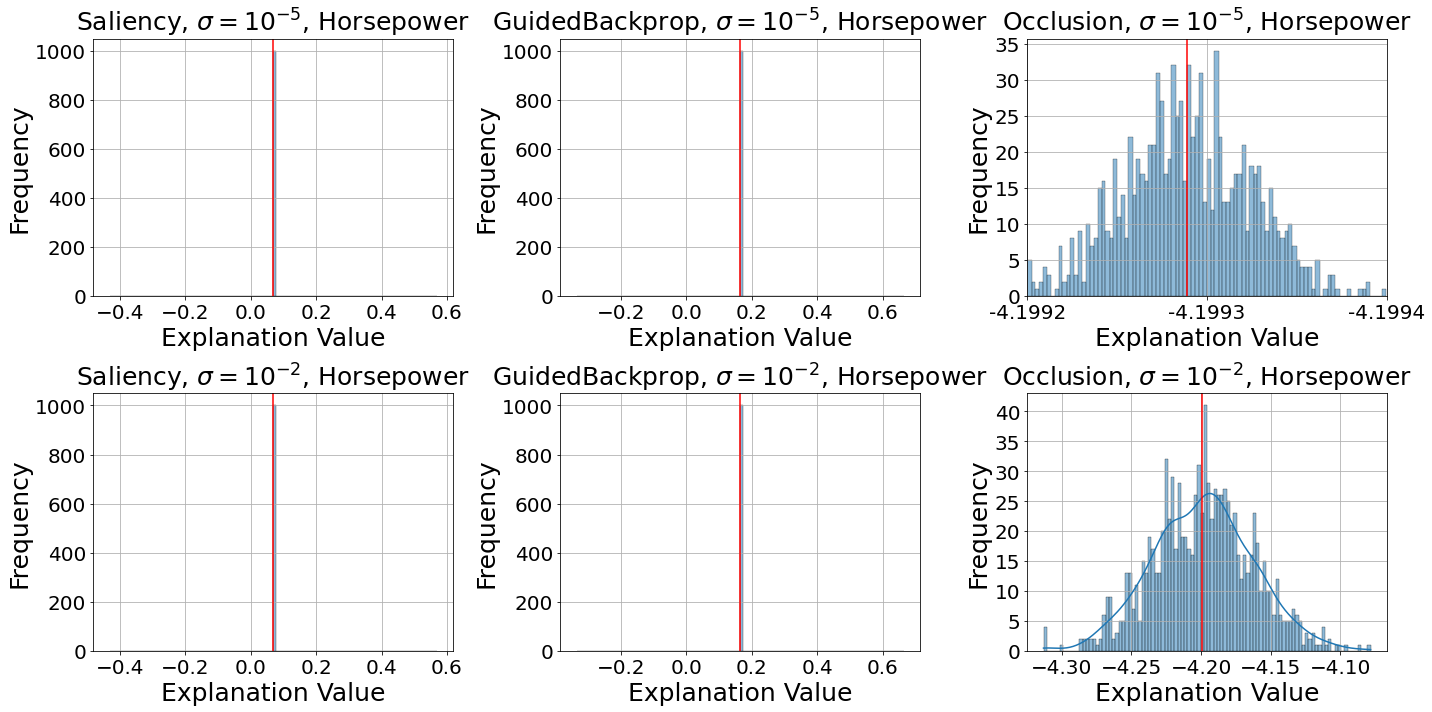}}
    \hfill
    \subfloat[MNIST (classification)]{\includegraphics[width=1.0\textwidth]{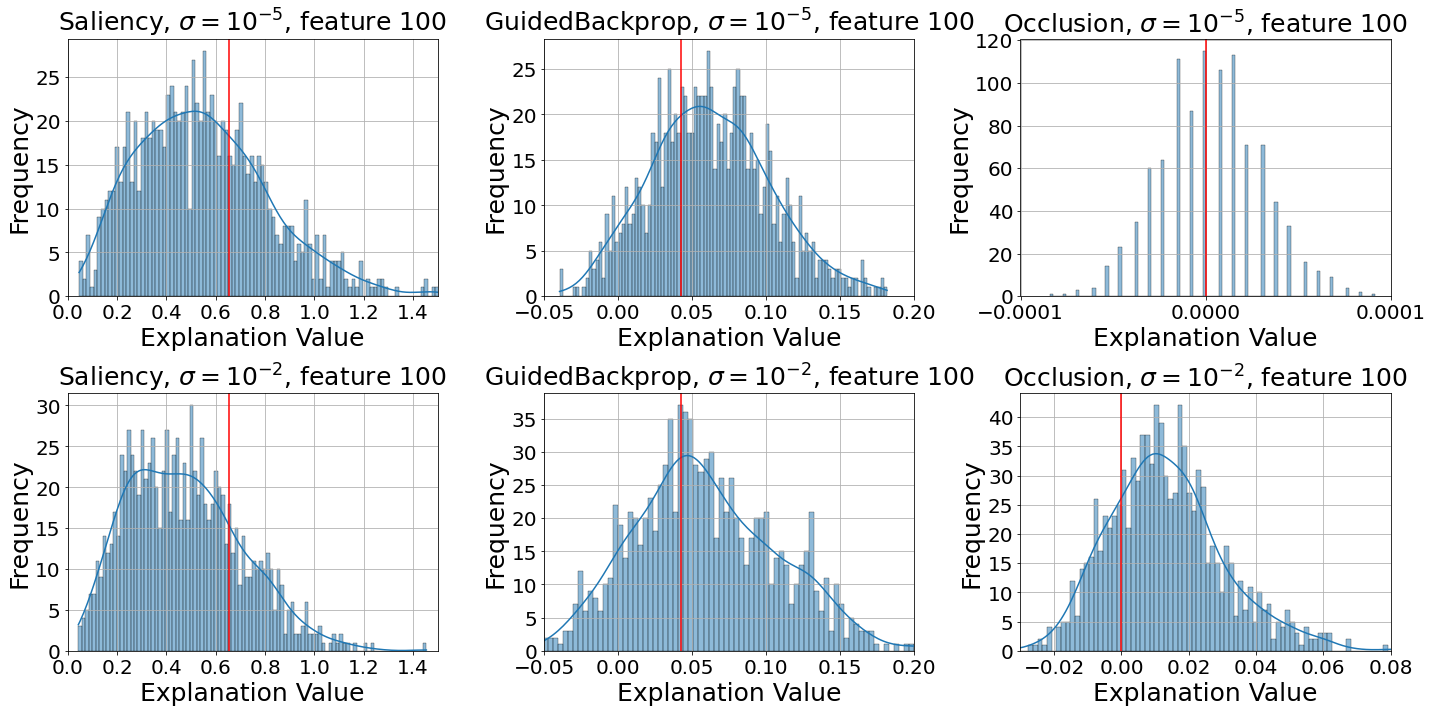}}
    \hfill
    \caption{Zero variance in explanations computed via input perturbations on Auto MPG for Saliency and GuidedBackprop regardless of $\sigma^2$ (\textbf{(a)}, first two columns, Case 2) vs proportionally increasing variance in explanations with $\sigma^2$ for Occlusion (\textbf{(a)}, third column, Case 1). Equally high variance in explanations computed via input perturbations on MNIST for Saliency and GuidedBackprop regardless of $\sigma^2$ (\textbf{(b)}, first two columns, Case 3) vs proportionally increasing variance in explanations with $\sigma^2$ for Occlusion (\textbf{(a)}, third column, Case 1). Also notice the approximate Gaussian shape of the distributions computed with Occlusion. The distributions were computed from 1000 random perturbations of a single randomly drawn sample in each dataset. The red line marks the value of the unperturbed explanation feature value; here, exemplarily displayed for the feature 'Horsepower' of the sample row in Auto MPG and the 100th pixel of the sample image in MNIST. Note that there is no curve fit computed for explanation variances near 0, e.g. \textbf{(a)}-row 1-column 3 and \textbf{(b)}-row 1-column 3.}
    \label{fig:input_normality}
\end{figure}

\begin{figure}
    \centering
    \subfloat[Auto MPG (regression)]{\includegraphics[width=1.0\textwidth]{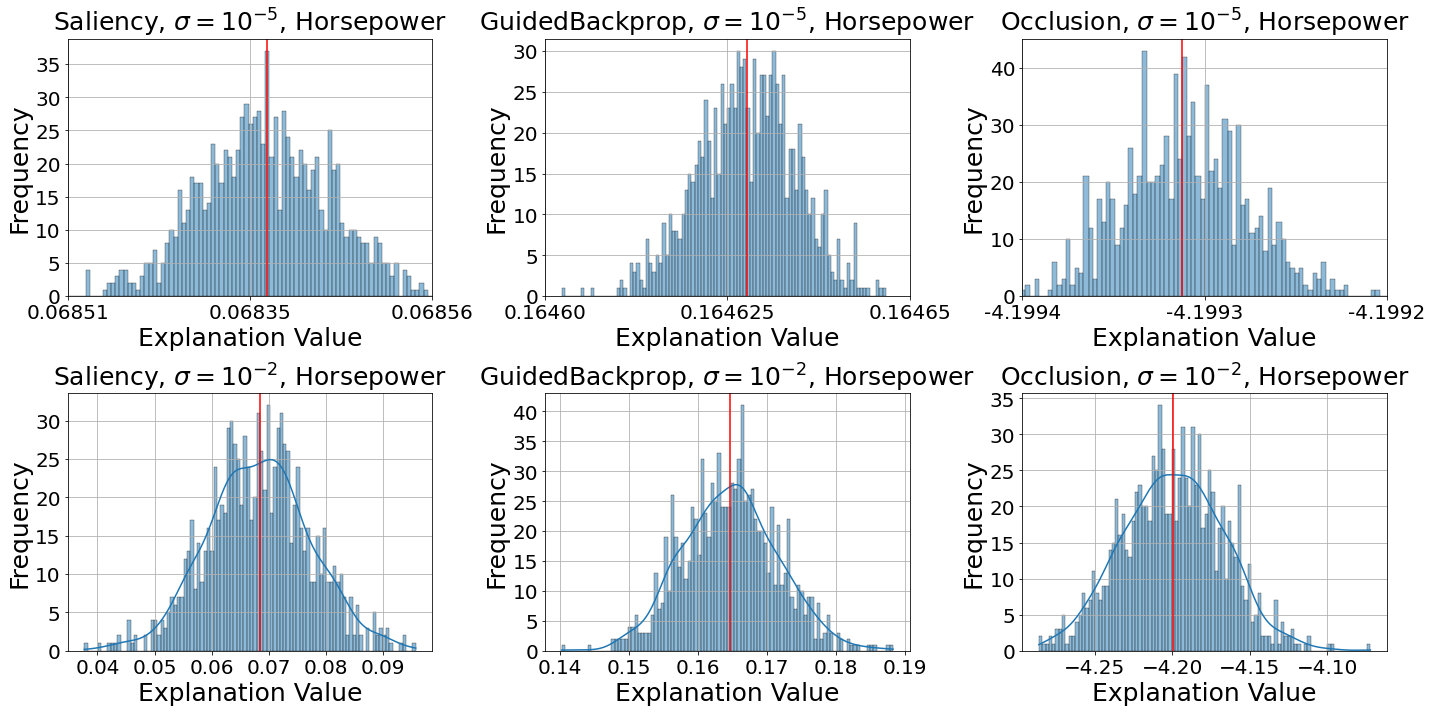}}
    \hfill
    \subfloat[MNIST (classification)]{\includegraphics[width=1.0\textwidth]{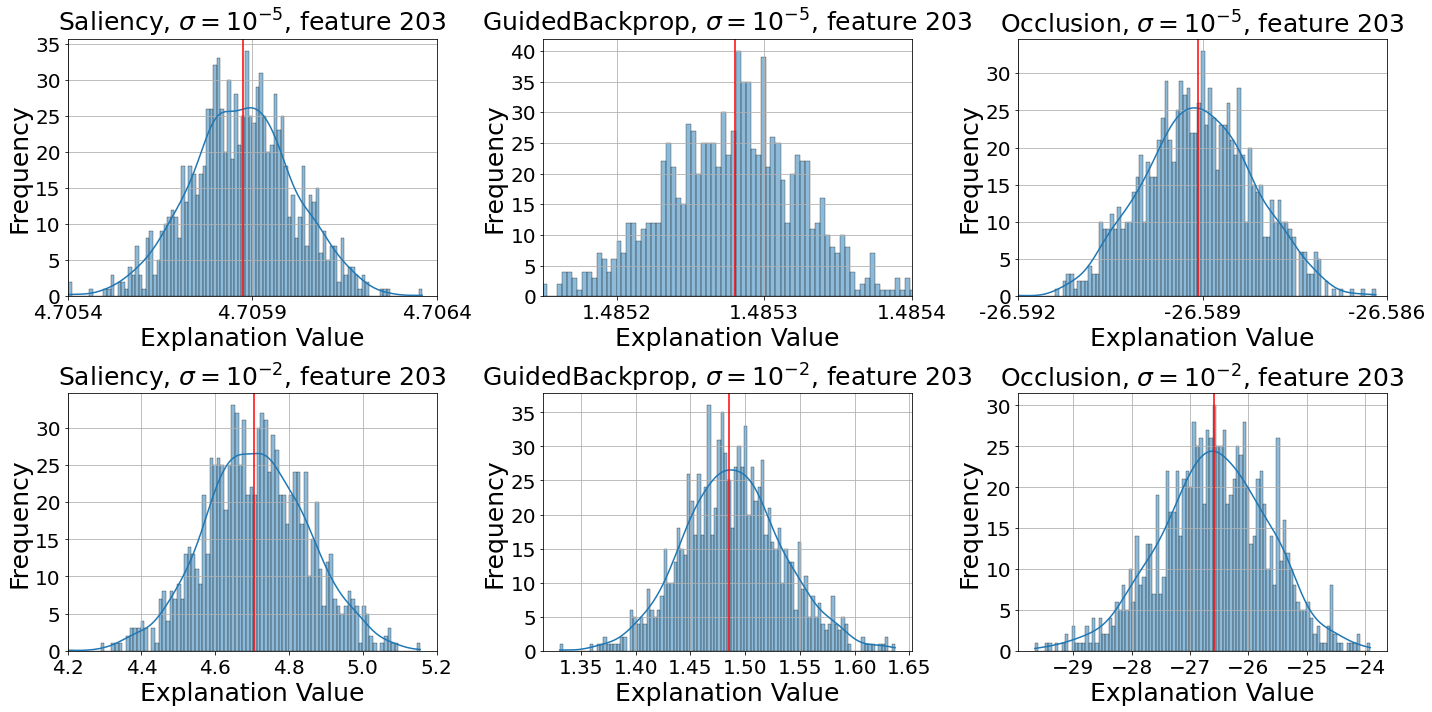}}
    \hfill
    \caption{Proportionally increasing variance in the explanations computed via model weights perturbations with increasing $\sigma^2$ across all XAI methods and tasks. Also notice the approximate Gaussian shape of the distributions, which were computed from 1000 random perturbations of a single randomly drawn sample in each dataset. The red line marks the value of the unperturbed explanation feature value; here, exemplarily displayed for the feature 'Horsepower' of the sample row in Auto MPG and the 203rd pixel of the sample image in MNIST. We chose a different pixel for visualization here as opposed to the one in \autoref{fig:input_normality} because many background explanation features computed by Occlusion are mapped to 0 and no real distribution can be plotted. Note that there is no curve fit computed for explanation variances near 0, e.g. \textbf{(a)}-row 1 and \textbf{(b)}-row 1-column 2.}
    \label{fig:weights_normality}
\end{figure}

In contrast to the first-order approximation assumption, we observe that for some methods, empirical uncertainties saturate below a certain perturbation scale rather than decreasing similarly to the linearized predictions. This is the case for the gradient-based methods applied on MNIST with perturbations in $x$ - \autoref{fig:uncert_input_agg}. The distributions shown in \autoref{fig:input_normality} also confirm that the variance in the perturbed explanations stays equally high regardless how low $\sigma^2$ is. Our further experiments revealed that by removing the ReLU activation in the CNN, this plateau in the gradient-based methods turns either into Case 1 (for GradientInput and Integrated Gradients) or Case 2 (for Saliency and GuidedBackprop). Apparently, the nonlinearity introduced by ReLU limits the sensitivity of explanations to input perturbations, causing a saturation effect. As a side note, methods like LIME \cite{salmaps4} and KernelSHAP \cite{kernelshap} also exhibit similar plateaus. However, in their cases, the plateau effect can be attributed to intrinsic randomness in their sampling-based estimations, introducing a form of irreducible noise that does not scale with $\sigma^2$. Since we narrowed the focus in this paper on input and model weights perturbations, analyzing the effect of explanation parameters $\theta$ is beyond the scope of this work.
\section{Discussion and Conclusion}\label{sec:conclusion}
Understanding the reliability of explanations is a fundamental prerequisite for responsible usage of ML technology. Previous work has repeatedly identified inconsistencies in popular XAI approaches\cite{Adebayo2020,salcritique2}. A key aspect that has not been explored extensively is how uncertainties in input data or model parameters impact uncertainties of explanations. In this work, we developed a model of uncertainty propagation in XAI and evaluated empirical as well as analytical estimators. In our experiments we find that XAI methods often fail to propagate uncertainties reliably. These results have implications for practical use cases in which uncertainties of explanations are relevant, but also for theoretical studies on the quality of ML explanations.
\subsection{Expanding to Other Models and Use Cases}

While our study focuses on feature attribution methods applied to neural networks in classification and regression tasks, the framework for uncertainty propagation in XAI is general and can be extended to a wide range of ML problem settings. The key components of our analysis do not depend on the specific architecture or task but rather on the ability to introduce controlled perturbations and estimate sensitivity, which we achieve via finite difference approximations. This is not only very similar to some of the most popular XAI methods, such as LIME \cite{salmaps4}, the approach is also model agnostic in that it is readily applicable beyond neural networks, including models like SVMs and decision trees, and to other tasks such as NLP or structured prediction. Other XAI methods - concept-based, example-based, counterfactuals - can also be analogously expressed as an explainer function $e_{\theta}(x, f)$ and integrated into our framework~\cite{uxai_review}. For more details we refer the interested reader to our publicly released code repository.

\subsection{Limitations and Extensions}

Our study focuses specifically on uncertainty stemming from input and model weight perturbations. However, as presented in \autoref{subsec:formal_uxai}, we are aware that uncertainty in explanations can also arise from the internal parameters of the explanation method itself. Methods such as LIME \cite{salmaps4} and KernelSHAP \cite{kernelshap} rely on stochastic sampling processes, introducing additional variance beyond what is captured in our framework. Extending our analysis to account for uncertainties inherent in explanation methods — such as the choice of hyperparameters, sampling strategies or surrogate model approximations — would provide a more comprehensive picture of uncertainty in XAI.

Our methodology relies on Gaussian perturbations to model uncertainty in both input data and model parameters. While Gaussian noise is a common choice in uncertainty quantification due to its mathematical tractability and connections to first-order approximations, it does not capture all forms of real-world uncertainty. For instance, real-world noise can be heteroscedastic, structured or follow skewed or heavy-tailed distributions that are not well approximated by a Gaussian assumption. Future research could extend our framework to non-Gaussian perturbations such as adversarial noise, multimodal distributions or dataset-specific perturbation models and provide a more comprehensive view of explanation robustness. Another aspect that remains to be explored in future work is an in-depth investigation on how different uncertainty models affect explanation variance and whether certain XAI methods exhibit different sensitivities to non-Gaussian uncertainties.

A key computational consideration when comparing empirical and analytical uncertainty estimators is the cost of computing the Jacobian of the explainer function $e$ via finite differences. If evaluating $e$ is computationally expensive, obtaining the Jacobian can become intractable or numerically unstable, especially for high-dimensional inputs. Established regularization schemes based on empirical (via cross-validation) or analytical shrinkage estimators could be useful in that setting \cite{ledoitWellconditionedEstimatorLargedimensional2004}. However, once the Jacobian is computed, the analytical approach provides uncertainty estimates for any perturbation scale $\sigma$ without requiring additional repeated perturbations and evaluations at each $\sigma$, which can be significantly more expensive in practice.

Beyond our specific experiments, our methodology provides a general framework for assessing uncertainty in explanations, which can be applied to a wide range of XAI techniques. By comparing empirical and analytical uncertainty measures, practitioners can evaluate whether an explanation method provides stable and reliable attributions under perturbations. This assessment can help guide decisions on whether to communicate uncertainty in explanations, and if so, in what form. For instance, if empirical and analytical uncertainties align well, users might confidently report uncertainty estimates; if they diverge significantly, it may indicate that the explanation method is unreliable under the given perturbation regime. Future work could explore how different applications tolerate uncertainty in explanations and determine appropriate thresholds for integrating UXAI into decision-making processes.

\section*{Acknowledgments} This research was supported by the Federal Ministry for Economic Affairs and Climate Action of Germany for the RIWWER project (with the project number 01MD22007H, 01MD22007C), the German Federal Ministry of Education and Research grant number 16SV8856 and 16SV8835, by the Einstein Center Digital Future, Berlin, and by the German Research Foundation (DFG) - Project number: 528483508 - FIP 12. The funder played no role in study design, data collection, analysis and interpretation of data, or the writing of this manuscript. 

\begin{credits}
\subsubsection{\discintname}
The authors have no competing interests to declare that are relevant to the content of this article.
\end{credits}
%
%

\bibliographystyle{splncs04}
\bibliography{references}

\newpage
\section*{Appendix}

\begin{algorithm}[H]
\caption{Analytical and Empirical UXAI w.r.t. Input Perturbation}\label{alg:inp_pert}
\begin{algorithmic}[1]
\State \textbf{Input:} Model $f$, input sample $x$, set of explainers $\mathcal{E}$, set of perturbation standard deviations $s$, number of perturbations $N$, differential step size $\delta$
\State \textbf{Output:} Analytical and empirical variance estimates for each explainer

\For{\textbf{each} explainer $e \in \mathcal{E}$}
    \State Compute explanation $e(x, f)$ for unperturbed input
    \State Initialize partial derivative matrix $\mathbb{J}_{e,x} \in \mathbb{R}^{m \times n}$

    \Comment{Compute Jacobian Block}
    \For{\textbf{each} feature index $i$ in $\{1, ..., n\}$}
        \State Compute differential step in input: $x' = x + \delta \cdot \mathbf{1}_i$
        \State Compute explanation $e(x', f)$
        \State Finite Differences for local derivatives: $\mathbb{J}_{e,x}[:, i] = \cfrac{e(x', f) - e(x, f)}{\delta} \approx \cfrac{\partial e}{\partial x_i}$
    \EndFor

    \Comment{Compute Analytical Covariance}
    \For{\textbf{each} $\sigma \in s$}
        \State Compute analytical covariance: $\Sigma_{\text{lin}} = \sigma^2 \cdot \mathbb{J}_{e,x} \cdot \mathbb{J}_{e,x}^T \in \R^{m \times m}$
        \State Analytical MUE($\sigma, x$): $\cfrac{\text{Tr}(\Sigma_{\text{lin}})}{m \cdot ||e(x,f)||_2^2}$
    \EndFor

    \Comment{Compute Empirical Covariance}
    \For{\textbf{each} $\sigma \in s$}
        \State Generate $N$ perturbed inputs: $\widetilde{x}_k = x + \Delta x_k, \Delta x_k \sim \mathcal{N}(0, \sigma^2 I), k = 1, \dots, N$
        \State Compute explanations: $e(\widetilde{x}_k, f), \quad k = 1, \dots, N$
        \State Compute empirical covariance: $\Sigma_{\text{MC}} = \text{Cov}(e(\widetilde{x}_1, f), \dots, e(\widetilde{x}_N, f))$
        \State Empirical MUE: $\cfrac{\text{Tr}(\Sigma_{\text{MC}})}{m \cdot ||e(x,f)||_2^2}$
    \EndFor
\EndFor
\end{algorithmic}
\end{algorithm}


\begin{figure}[hbt!]
    \centering
    \subfloat[Low Variance Regime]{\includegraphics[width=1.0\textwidth]{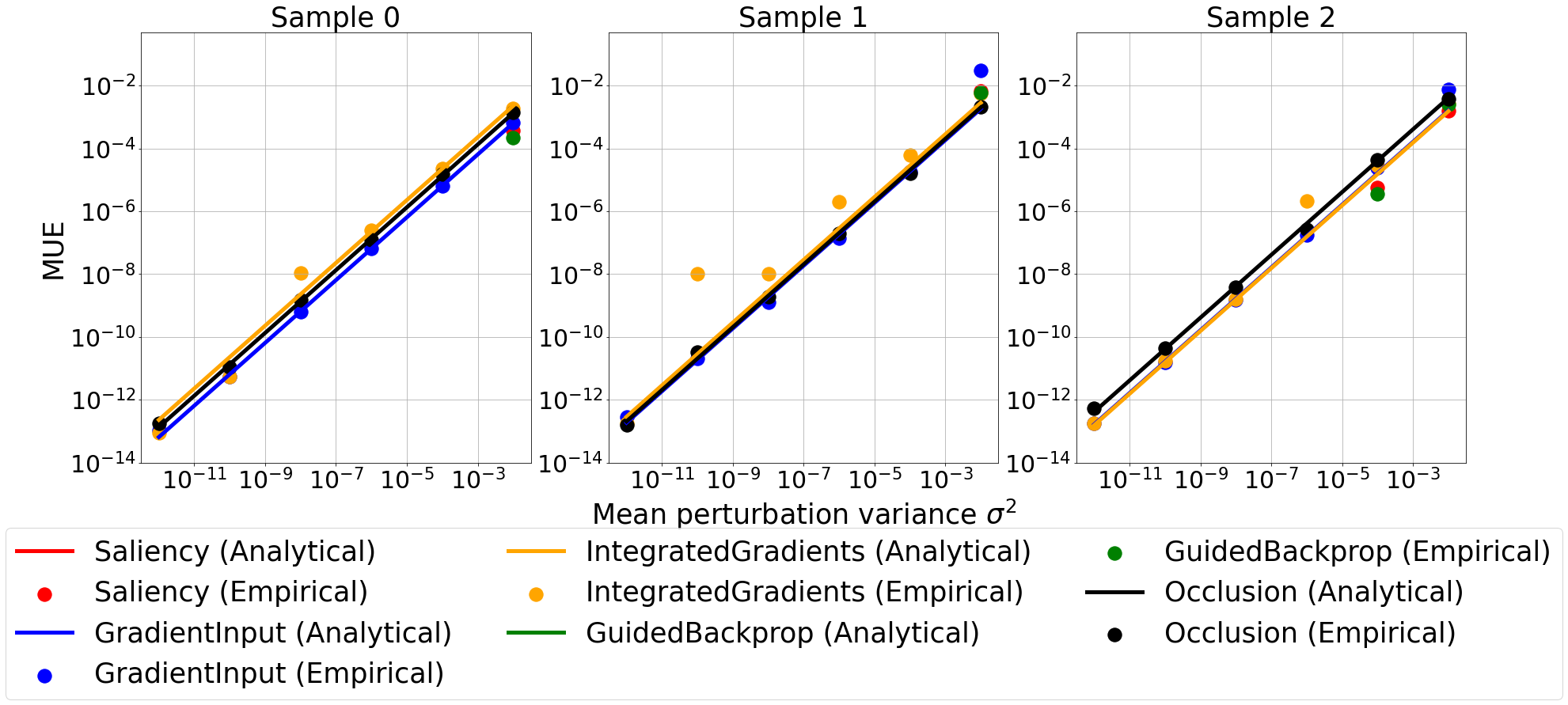}}
    \hfill
    \subfloat[High Variance Regime]{\includegraphics[width=1.0\textwidth]{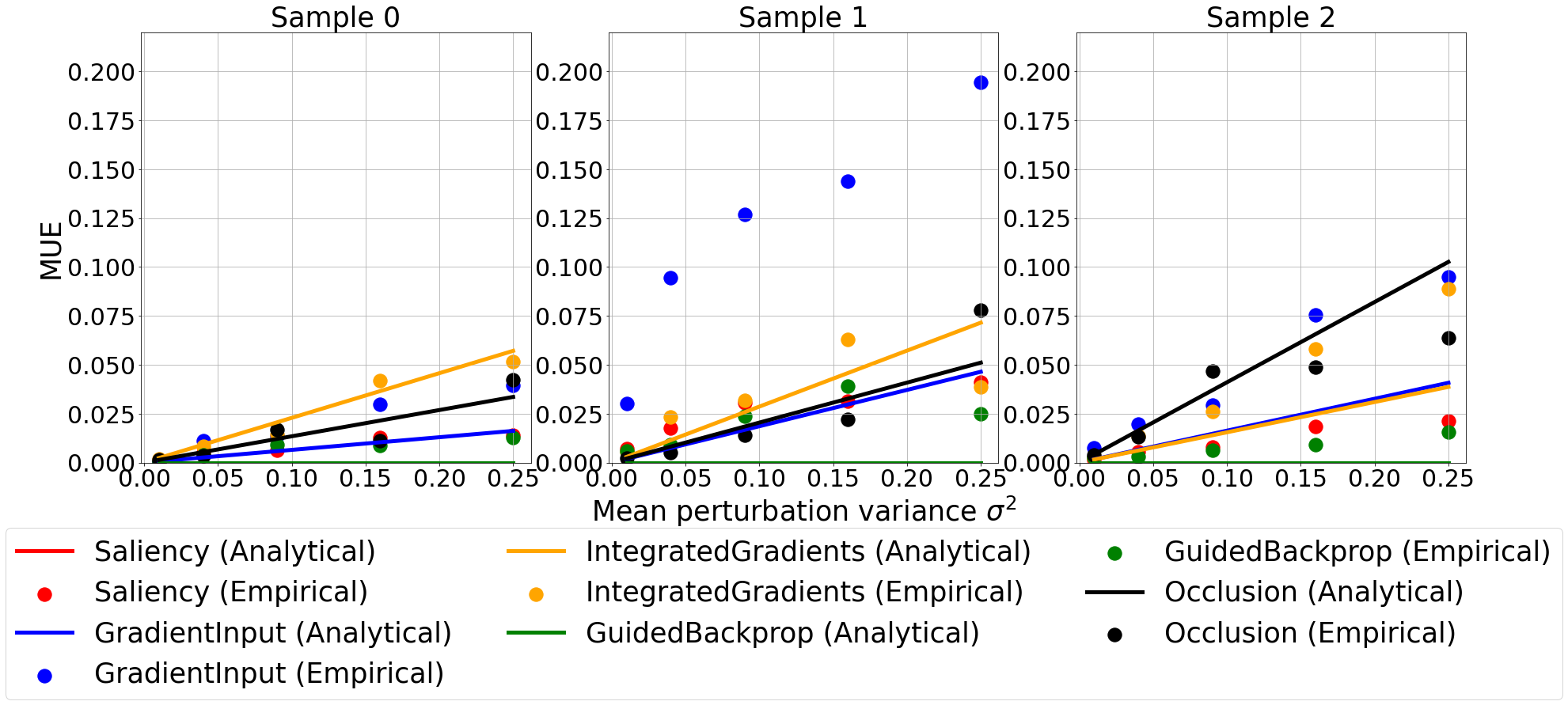}}
    \hfill
    \caption{Examples of input perturbations on Auto MPG. Notice the good linear fit of the analytical MUE line to the empirical MUE dots for GradientInput, Integrated Gradients and Occlusion - Case 1. The linearizations for Saliency and GuidedBackprop are 0 and not visible in the plots - Case 2. The continuous lines show the MUE forecast computed from $\Sigma_{\text{lin}}$, while the dots mark the MUE computed from $\Sigma_{\text{MC}}$. Please beware of the different axes scales (logarithmic in \textbf{(a)}, linear in \textbf{(b)}).}
    \label{fig:uncert_input_ex_tab}
\end{figure}

\begin{figure}[hbt!]
    \centering
    \subfloat[Low Variance Regime]{\includegraphics[width=1.0\textwidth]{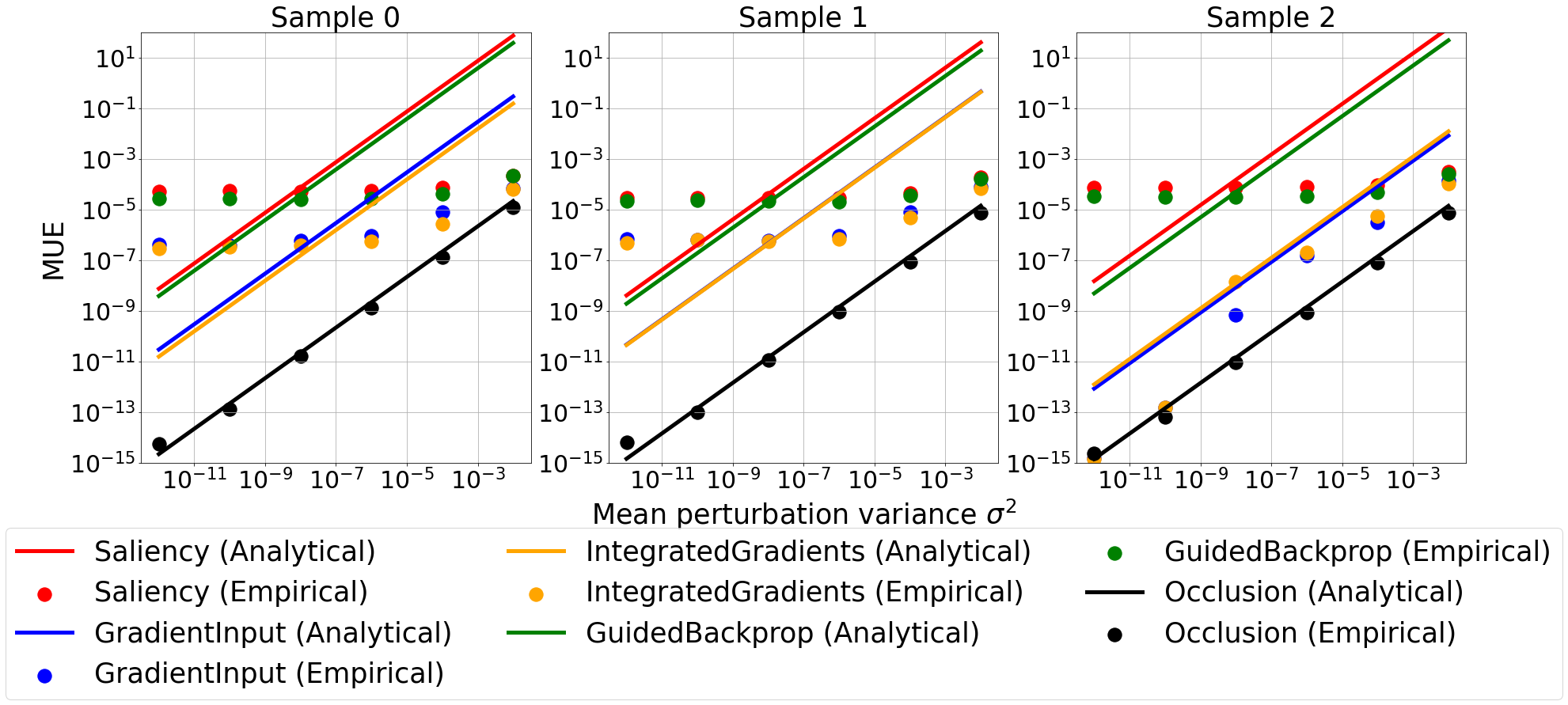}}
    \hfill
    \subfloat[High Variance Regime]{\includegraphics[width=1.0\textwidth]{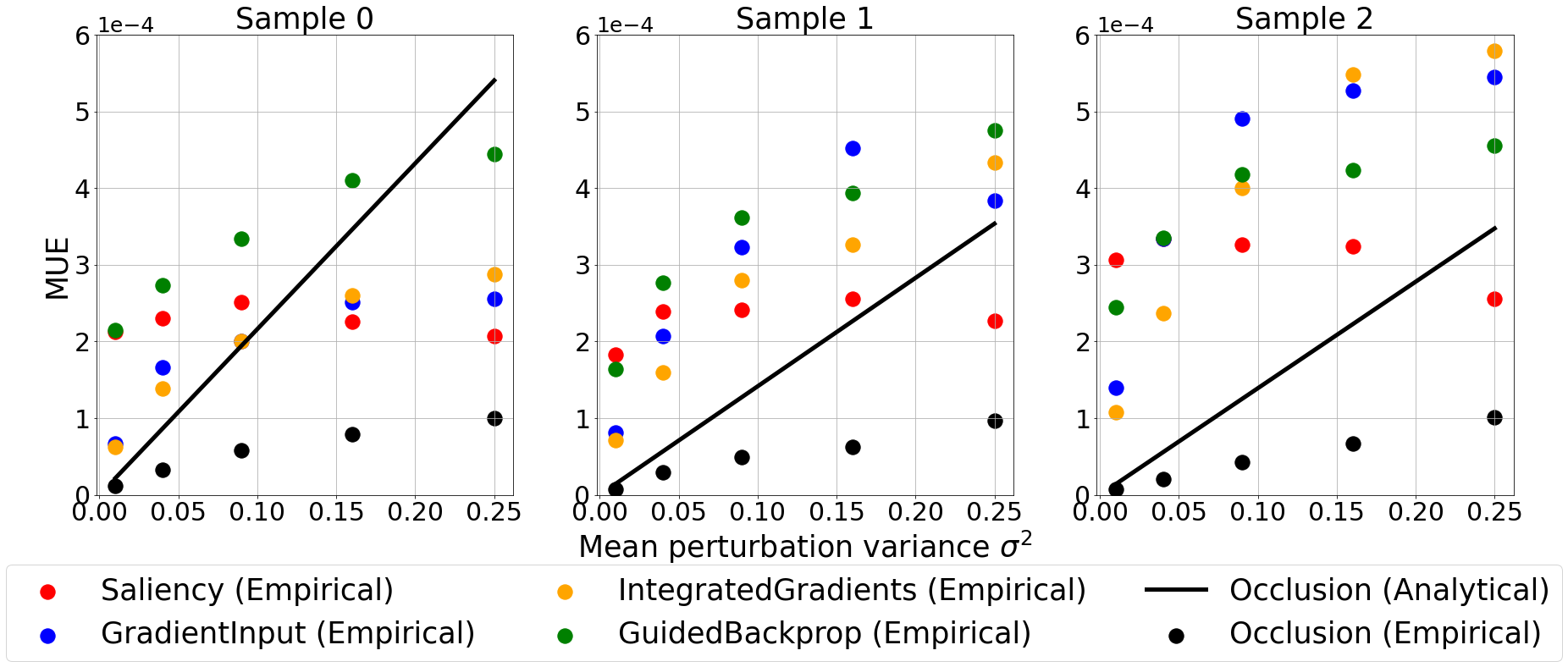}}
    \hfill
    \caption{Examples of input perturbations on MNIST. Notice the good linear fit for Occlusion - Case 1, and the plateaus for the gradient-based methods in \textbf{(a)} - Case 3. The continuous lines show the MUE forecast computed from $\Sigma_{\text{lin}}$, while the dots mark the MUE computed from $\Sigma_{\text{MC}}$. Please beware of the different axes scales (logarithmic in \textbf{(a)}, linear in \textbf{(b)} scaled by $10^{-4}$). Lines not visible in the high variance regime plots are left out because of their high slope.}
    \label{fig:uncert_input_ex_mnist}
\end{figure}

\begin{figure}[hbt!]
    \centering
    \subfloat[Low Variance Regime]{\includegraphics[width=1.0\textwidth]{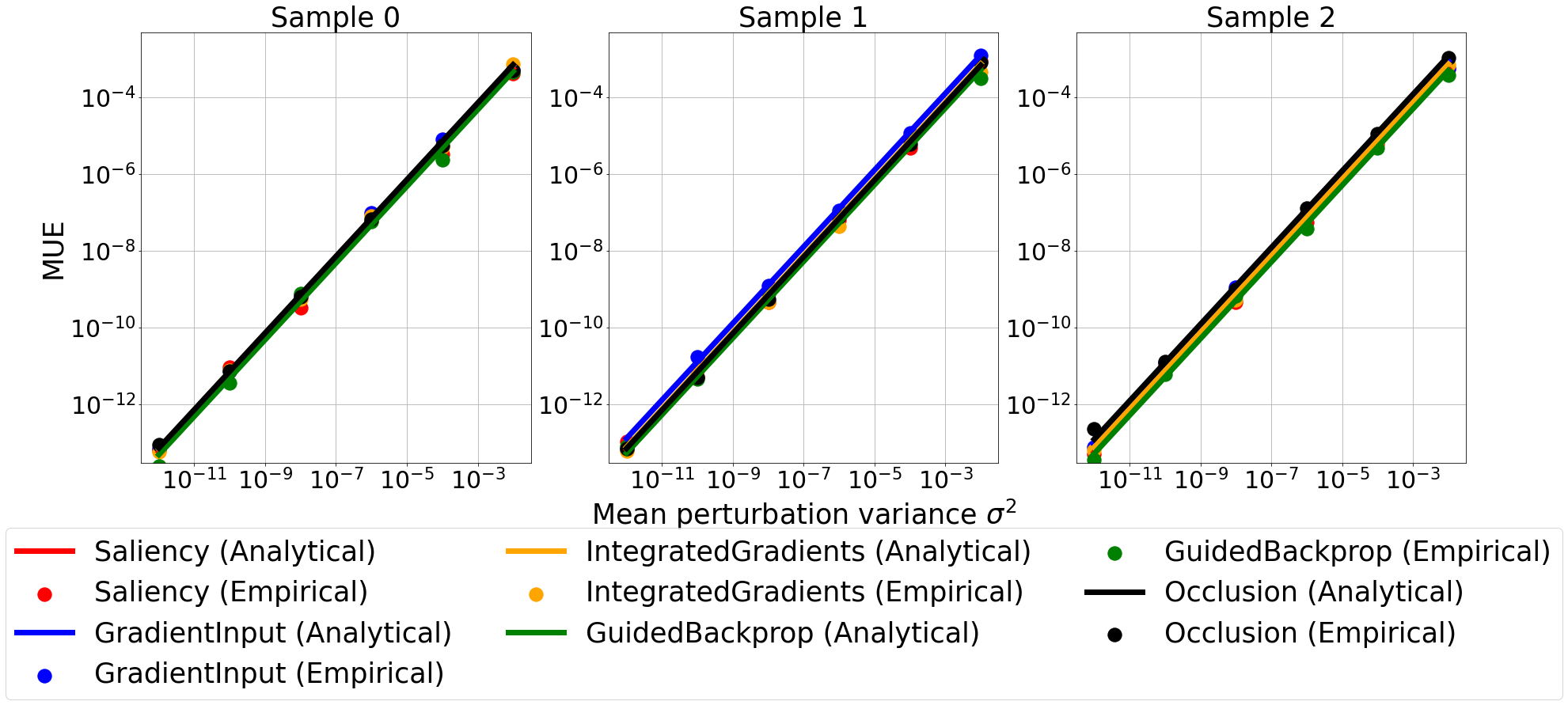}}
    \hfill
    \subfloat[High Variance Regime]{\includegraphics[width=1.0\textwidth]{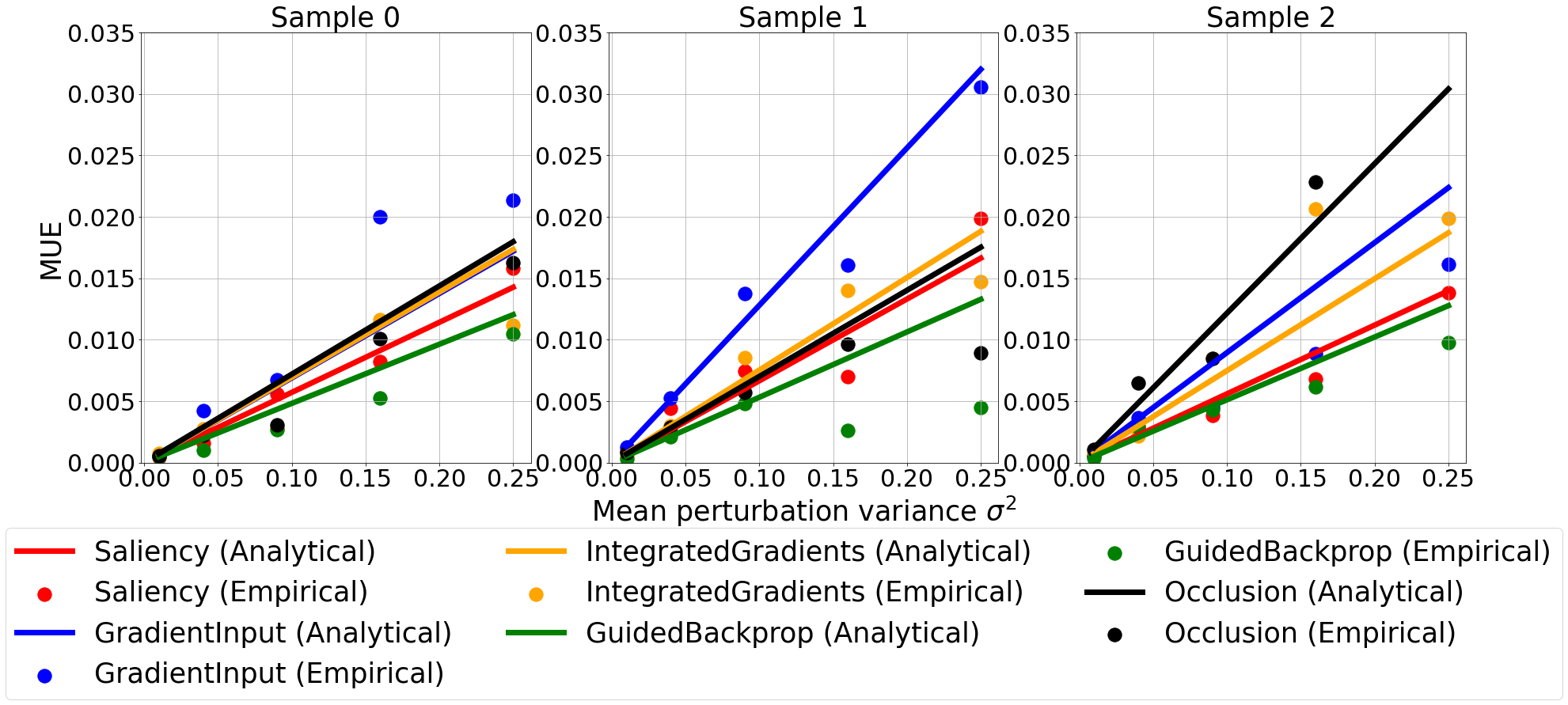}}
    \hfill
    \caption{Examples of model weights perturbations on Auto MPG. Notice the very good linear fit across all XAI methods and variance regimes - Case 1. The continuous lines show the MUE forecast computed from $\Sigma_{\text{lin}}$, while the dots mark the MUE computed from $\Sigma_{\text{MC}}$. Please beware of the different axes scales (logarithmic in \textbf{(a)}, linear in \textbf{(b)}).}
    \label{fig:uncert_weights_ex_tab}
\end{figure}

\begin{figure}[hbt!]
    \centering
    \subfloat[Low Variance Regime]{\includegraphics[width=1.0\textwidth]{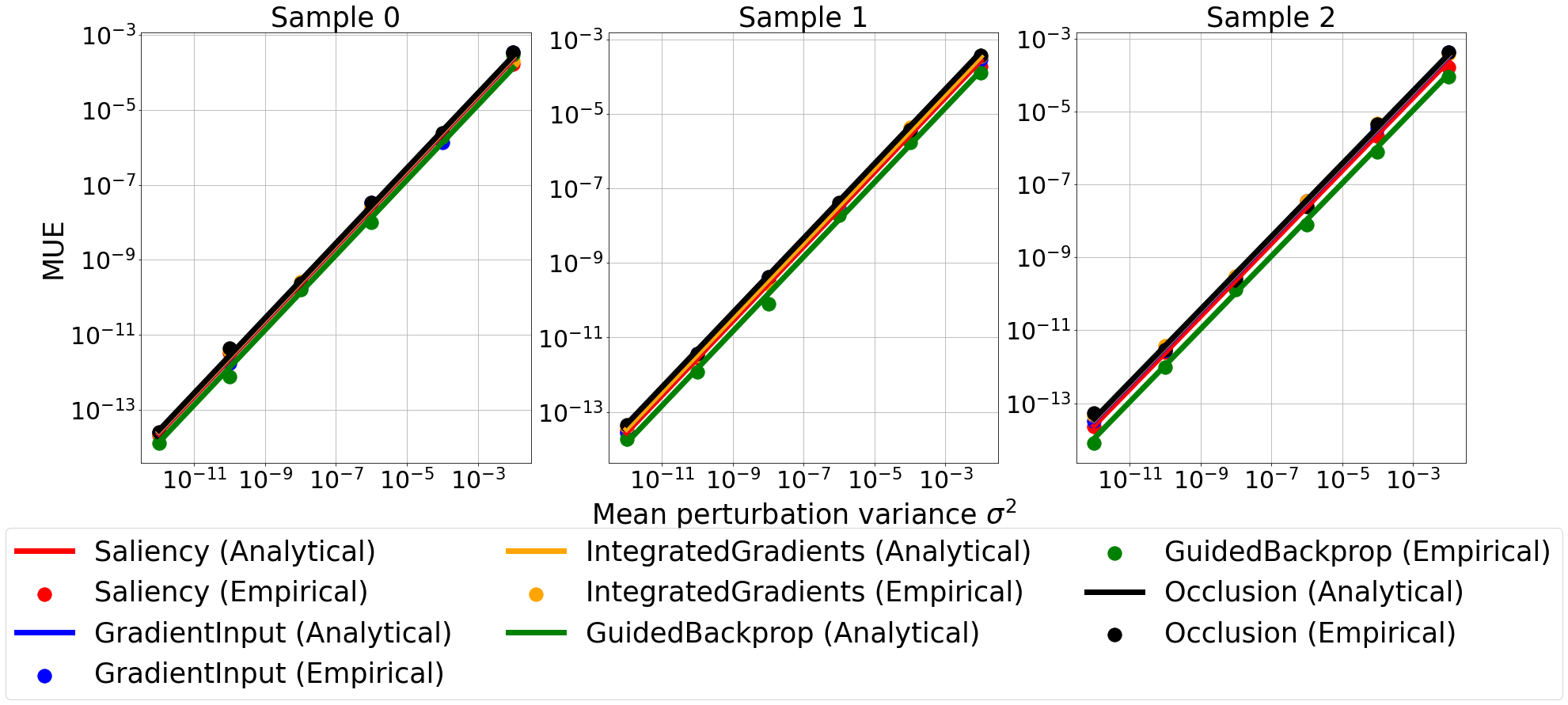}}
    \hfill
    \subfloat[High Variance Regime]{\includegraphics[width=1.0\textwidth]{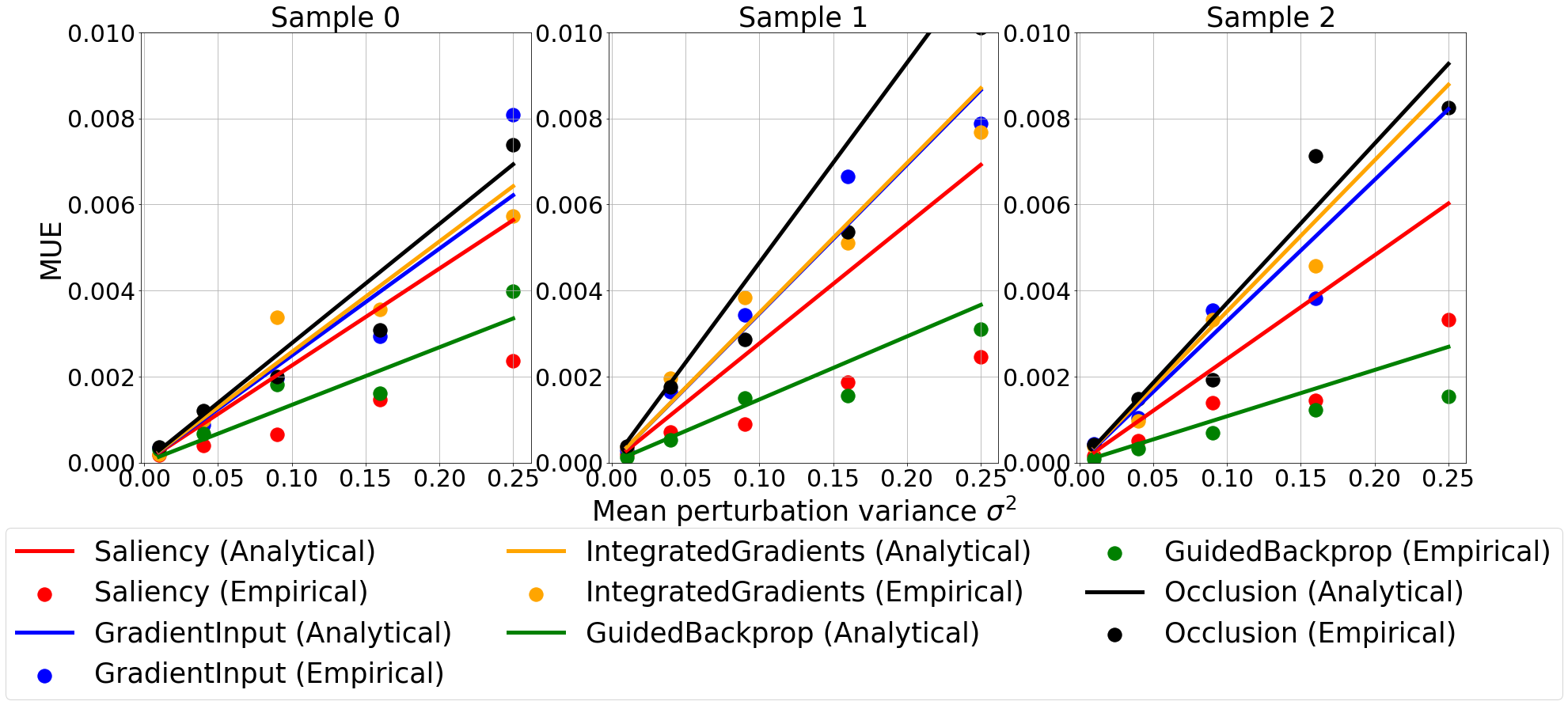}}
    \hfill
    \caption{Examples of model weights perturbations on MNIST. Notice the very good linear fit across all XAI methods and variance regimes - Case 1. The continuous lines show the MUE forecast computed from $\Sigma_{\text{lin}}$, while the dots mark the MUE computed from $\Sigma_{\text{MC}}$. Please beware of the different axes scales (logarithmic in \textbf{(a)}, linear in \textbf{(b)}).}
    \label{fig:uncert_weights_ex_mnist}
\end{figure}


\begin{figure}
    \centering
    \subfloat[Diagonal of $\mathbb{J}_{e,x}$]{\includegraphics[width=1.0\textwidth]{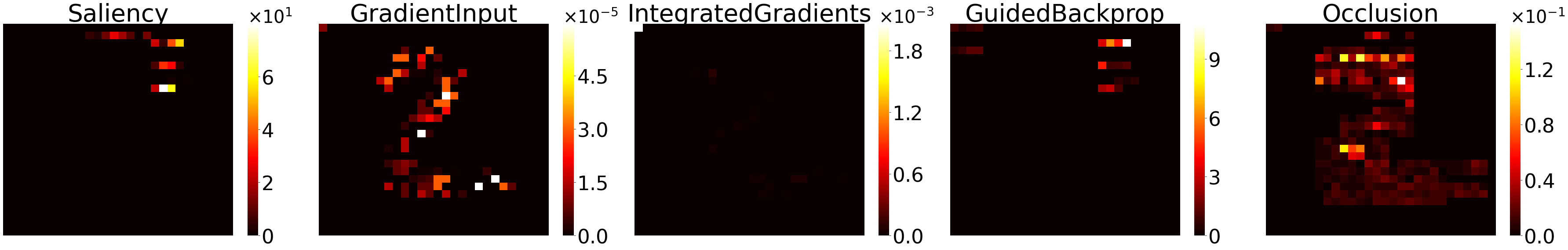}}
    \hfill
    \subfloat[Diagonal of $\Sigma_{\text{lin}}$]{\includegraphics[width=1.0\textwidth]{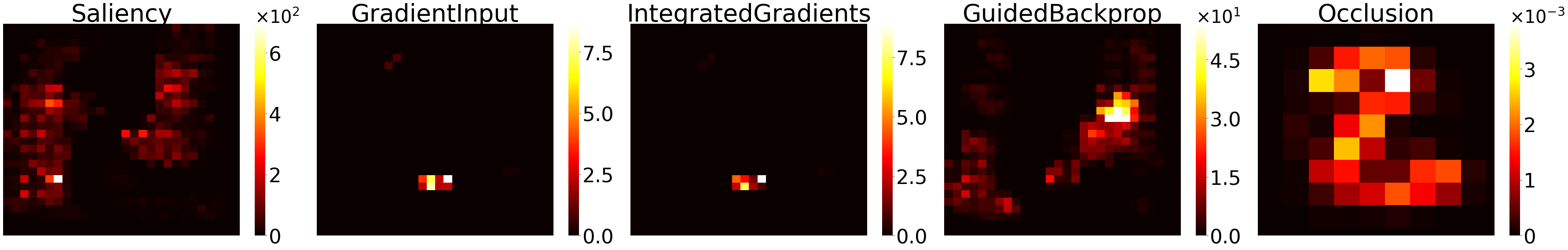}}
    \hfill
    \subfloat[Diagonal of $\Sigma_{\text{MC}}$]{\includegraphics[width=1.0\textwidth]{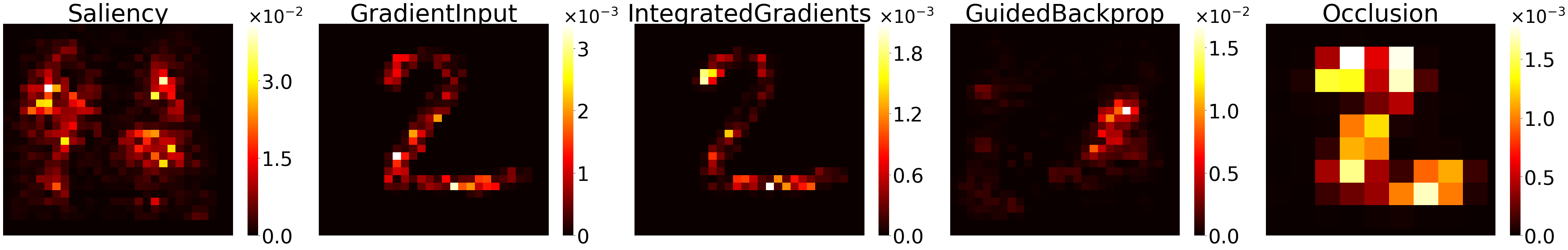}}
    \hfill
    \caption{Possible visualizations of the uncertainty in the explainer $e$ w.r.t. the uncertainty in the input $x$ on MNIST. By plotting the diagonal of the partial derivative $\mathbb{J}_{e,x}$ or the covariance matrices $\Sigma_{\text{lin}}$ and $\Sigma_{\text{MC}}$, we can get a glimpse into which original pixels in the unperturbed image $x$ have a high effect on the corresponding explanation feature at the same position $(i, j)$.}
    \label{fig:diagonals}
\end{figure}

\end{document}